\documentclass[11pt,a4paper]{article}
\usepackage[hyperref]{acl2021}
\usepackage{times}
\usepackage{latexsym}

\usepackage{microtype}

\aclfinalcopy 
\def\aclpaperid{3215} 

\usepackage{multirow}

\usepackage{booktabs}
\usepackage[normalem]{ulem} 
\usepackage[utf8]{inputenc}
\usepackage[T1]{fontenc}
\usepackage[noabbrev,capitalize]{cleveref}
\usepackage[disable]{todonotes}
\newcommand{\jp}[2][inline]{\space\todo[color=green!60,#1,size=\tiny]{{JP: #2}}}
\newcommand{\jj}[2][inline]{\space\todo[color=red!60,#1,size=\tiny]{{JOSEF: #2}}}
\newcommand{\dv}[2][inline]{\space\todo[color=blue!60,#1,size=\tiny]{{DUSAN: #2}}}
\newcommand{\mn}[2][inline]{\space\todo[color=magenta!60,#1,size=\tiny]{{MISO: #2}}}

\def\XXX#1{{\textcolor{red}{XXX #1}}}

 \definecolor{darkgreen}{rgb}{0.0, 0.7, 0.1}

\title{End-to-End Lexically Constrained Machine Translation \\  for Morphologically Rich Languages}

\author{Josef Jon \and João Paulo Aires \and Dušan Variš \and Ondřej Bojar \\
       Charles University\\
  \texttt{\{jon,aires,varis,bojar\}@ufal.mff.cuni.cz}  }

\date{}

\begin{document}
\maketitle
\begin{abstract}
Lexically constrained machine translation allows the user to manipulate the output sentence by enforcing the presence or absence of certain words and phrases.
Although current approaches can enforce terms to appear in the translation, they often struggle to make the constraint word form agree with the rest of the generated output.
Our manual analysis shows that 46\% of the errors in the output of a baseline constrained model for English to Czech translation are related to agreement.
We investigate mechanisms to allow neural machine translation to infer the correct word inflection given lemmatized constraints.
In particular, we focus on methods based on training the model with constraints provided as part of the input sequence.
Our experiments on the English-Czech language pair show that this approach improves the translation of constrained terms in both automatic and manual evaluation by reducing errors in agreement.
Our approach thus eliminates inflection errors, without introducing new errors or decreasing the overall quality of the translation.
\end{abstract}



\section{Introduction}
\label{sec:introduction}

In Neural Machine Translation (NMT), lexical constraining ~\cite{song-etal-2019-code, hokamp-liu-2017-lexically, post-vilar-2018-fast} involves changing the translation process in a way that desired terms appear in the model's output.
Translation constraints are useful in domain adaptation, interactive machine translation or named entities translation.
Current approaches focus either on manipulating beam search decoding ~\cite{hokamp-liu-2017-lexically, post-vilar-2018-fast, hu-etal-2019-improved} or training an NMT model using constraints alongside the input~\cite{dinu-etal-2019-training,song-etal-2019-code, chen-etal-2020-lexical}.

In inflected languages, constraints from both source and target sides may appear in numerous surface forms, which may result in errors during translation.
By enforcing the presence of a certain exact term on the target side, existing approaches fail to deal with word inflections.
As we show, they preserve the surface form of the word provided as constraint regardless of the context.
Morphologically rich languages have multiple forms of each word, e.g. inflections to nouns.
For satisfactory results in these languages, the constraint processing method needs to be capable of detecting any surface form on the source side and generating the correct surface form on the target side.

\begin{figure}[t]
    \centering
    \includegraphics[width=\columnwidth]{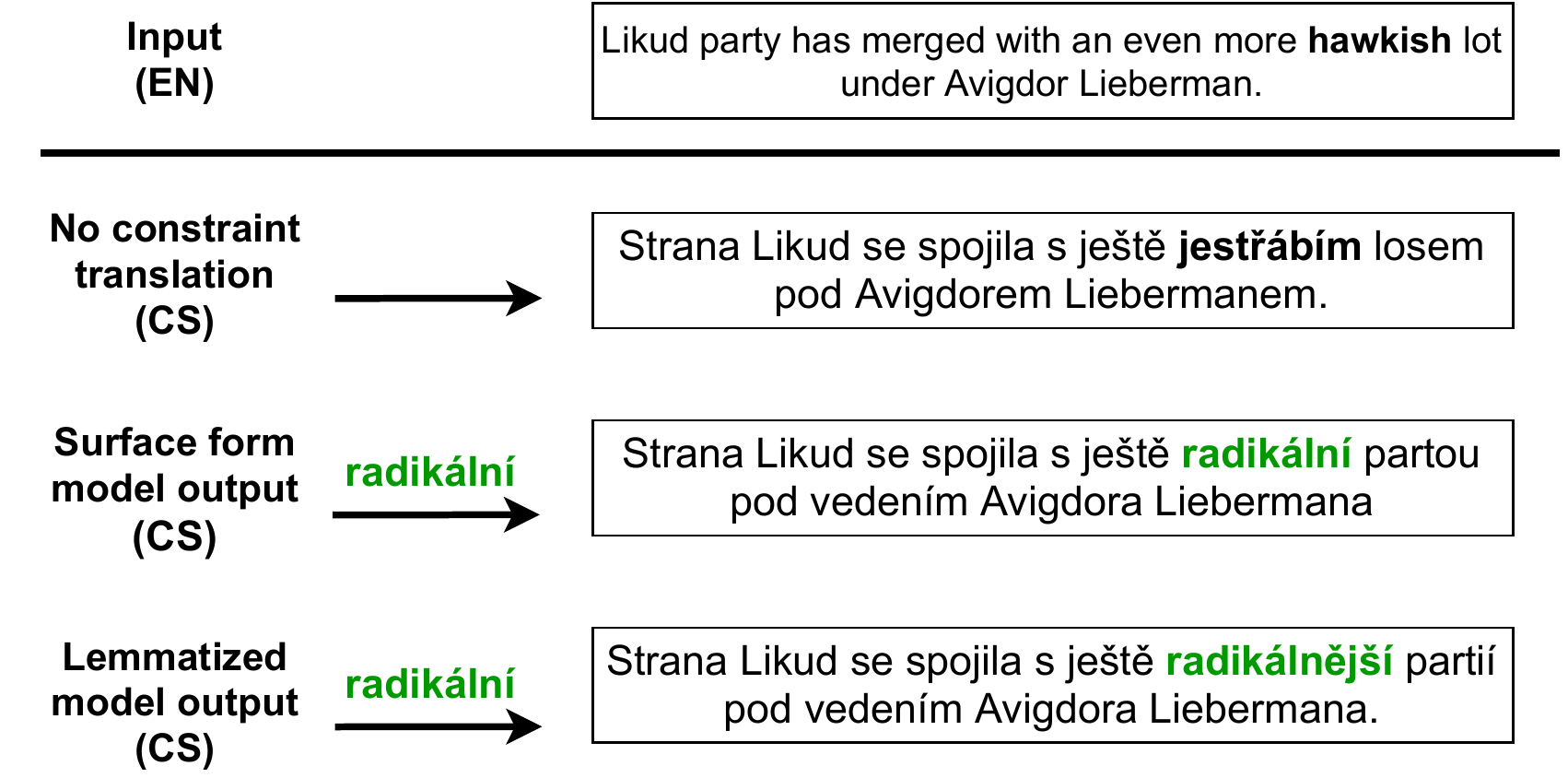}
    \caption{Comparison between constrained translations from English to Czech.}
    \label{fig:constrained_translation}
\end{figure}

To illustrate the problem, \cref{fig:constrained_translation} shows a sentence translation from English to Czech with outputs from three methods.
The first one is a no-constraint translation where ``hawkish'' is translated as ``jestřábím'' (literally ``hawkish'', no figurative meaning; followed by a further mis-translation of ``lot'').
The second is a constrained model requested to use the word form ``radikální'' (``radical'') in the output.
The constraint was satisfied but the adjective should have taken the comparative degree to match
the rest of the translation.
The third output is the result of a model that processes the input along with the canonical form constraint (``radikální'') and modifies the constraint inflection in the final translation (``radikálnější'') to correctly express the comparative form (although the translation of ``lot'' is worse than in previous case).



We evaluate different methods of lexically constrained machine translation on the Czech language.
We propose an approach to deal with word inflection in lexically constrained translation.
By training a model that receives lemmatized target constraints as the input alongside the source sentence, we improve the generation of constraints in forms matching the output context. 
We run experiments on both synthetic and real-world test scenarios.

\section{Related work}
\label{sec:const_translation}

In MT, there are scenarios where words that should or should not appear in the output are known upfront.
Common use cases include integration of domain-specific terminology and translation of named entities or rare words using a dictionary.
Such functionality was previously implemented in phrase-based systems \citep{okuma:dicts-in-pbmt:2008}, like Moses \citep{koehn-etal-2007-moses}.
In NMT, this task is not yet definitely solved, since the translation process is hard to interpret and influence.
 
\subsection{Output post-processing}
\label{subsec:output_post-processing with the section naming in the rest of the paper (only first letter uppercase?}

In order to enforce the presence of specific terms, some approaches post-process the output.
Prior to subword handling \citep{sennrich-etal-2016-neural,kudo-richardson-2018-sentencepiece}, unknown words were corrected by replacing them with word translation pairs from a bilingual dictionary
\citep{luong-etal-2015-addressing}. 
\citet{crego2016systrans} use placeholders to translate numbers and named entities.
Placeholders have also been found useful for translation of text with formal mark-up and its interaction with content \citep{hanneman-dinu:2020:WMT}.

\subsection{Constrained decoding}
\label{subsec:const_decoding}

An alternative way of adding constraints to the final translation is by manipulating the beam search decoding process.
\citet{anderson-etal-2017-guided} use a finite state machine (FSM) that recognizes target sentence with constraint patterns.
Each state of the FSM has its own beam and only hypotheses in beams that are in accepting states can be finished.
\citet{hasler-etal-2018-neural} improve upon this work by utilizing encoder-decoder attention weights to guide the placement of a constraint. 
\citet{chatterjee-etal-2017-guiding} also use attention weights and beam search look-ahead to choose constraint positions.

\citet{hokamp-liu-2017-lexically} present Grid Beam Search, which extends the usual beam search~\cite{och-ney-2004-alignment} with a mechanism to ensure the coverage of all constrains.
\citet{post-vilar-2018-fast} propose a similar but more efficient algorithm.
By dynamically reallocating the beam capacity, an arbitrary number of constraints can be processed within a constant width of the beam.

One shortcoming of the above methods is the slower inference compared to unmodified beam search models.
This issue is in large part solved by effective vectorized beam allocation~\cite{hu-etal-2019-improved}.
Another drawback of constrained decoding is a less fluent output, especially in morphologically rich languages, since we force the output to contain a phrase that may not be in agreement with the rest of the output.


\subsection{Learned constraining}

One way of integrating constraints into NMT is to provide them alongside the input sentence and train the model to be biased towards utilizing them.
This gives the user less direct control over the output translation and requires specially trained models.
On the other hand, these approaches are simple to implement, do not incur inference slowdown, and make the translation more robust in case of wrongly chosen constraints.
NMT models are often able to produce very fluent output \citep{popel-etal-2020-cubbitt}, making them capable to cope with inflections properly.
Thus, using this capability may yield better results than constrained decoding with heuristics for inflections in inflected languages.

\citet{dinu-etal-2019-training} use input factors to annotate source sentences with desired translations and train the model to copy these translations into the output sequence.
\citet{chen-etal-2020-lexical} append constraints to the end of the source sentence.
Their goal is to train the model to place constraints in the output translation without the need of a bilingual dictionary or a specified word alignment. 
\citet{song-etal-2019-code} also propose a data augmentation approach that uses constraints along the source as input during the model training.
Concurrently to our work, \citet{bergmanis-pinnis-2021-facilitating} modify \citet{dinu-etal-2019-training} approach by providing lemmatized word factors associated to random tokens in the source sentence.
With the lemmatized factors, they force the model to learn the correct inflection of the word in the translation.

The main difference between our work and most of the existing approaches is the use of lemmatized constraints to allow the model to correctly inflect them to agree with the output context. The concurrent work by \citet{bergmanis-pinnis-2021-facilitating} presents a very similar idea. They also use lemmatized forms of the constraints and let the model itself to generate correct surface form. While their choice of languages (English to Latvian) and their experimental setup was slightly different, the overall conclusions of their work agree with ours. 
The main difference is the approach to integration of the constraints. \citet{bergmanis-pinnis-2021-facilitating}  use factors to directly annotate to the source tokens with lemmas of their desired translations. We experimented with this approach (see \ref{app:factors}), but in most of the experiments, we opted for a simpler integration method, by concatenating desired target lemmas to the source sentence. This simplifies preparation of the training data by removing the need for source to target word alignment and as we show, hurts the performance only by a very slight margin.


\section{Proposed methods} 

Building upon the described techniques, we focus on allowing the model to choose the correct word form.
Our approaches are based on learned constraining, where the constraints are lemmatized during both training and test time. 

\subsection{Learned constraining}


In our approach, we append the target constraints as a suffix of the input sentences, same as ~\citet{chen-etal-2020-lexical}.
We use \texttt{<sep>} token to separate constraints from the input sentence, and \texttt{<c>} token to separate constraints from each other.
Inspired by~\citet{chen-etal-2020-lexical}, we shift the positional embeddings by 1024 for the constraint tokens. However, while \citet{chen-etal-2020-lexical} start each constraint on the same position, we shift the start of the constraint string and continue monotonically from there. We do not use any other techniques described in their work.
The following example illustrates an input to our baseline constrained model, passing two constraints (``plánováno'' and ``obcích'') along with the source text.
In this case, both constraints are in correct target surface forms, which are obtained from the reference translation.
Without knowledge of the reference, it is necessary to solve the problem of agreement of the constraint with the rest of the translation, which is the main goal of our work.

    \hspace*{-0.4cm}\begin{tabular}{p{7.4cm}}
    \\
        Source: Price increase is planned  mainly in larger municipalities. \texttt{<sep>} \textbf{plánováno} \texttt{<c>} \textbf{obcích}\\ \\
        
        Reference: Zvýšení cen je \textbf{plánováno} především ve větších \textbf{obcích}.\\ \\
    
    \end{tabular}
We also experimented with the factored translation approach introduced by~\citet{dinu-etal-2019-training} as a second constraint integration method.
In \cref{app:b}, we present the description of the method and a comparison with appending the constraints as a suffix.

\subsection{Preparing synthetic constraints}

To our current knowledge, there is no English-Czech dataset with provided constraints. Thus, we generate constraints from the existing parallel data.
We consider two approaches to generate constraints for the training and test data.
\vspace{-0.2cm}
\paragraph{Training}
The simplest method of obtaining target-side constraints is sampling random token subsequences from the reference sentence.
In our experiments, every token in the sentence can become a start of a constraint with probability of 0.3.
An open constraint finishes on each subsequent token with probability of 0.85 and multiple constraints for a single sentence are permitted (without overlapping). We did not optimize these probabilities, further gains may be obtained by a search for better values.
The constraint order is randomly permuted, since during the test time,  order of constraints in the target is not known beforehand. 
The second approach makes use of either a bilingual dictionary or a terminology database. If a translation pair from the dictionary is found in the source and target sentences, its target side can serve as the constraint. By this method, we also obtain alignment of the source and target expressions, which is useful for the factored translation approach (see \cref{app:factors}).

 \paragraph{Test time}
Given an input sentence and no reference translation, we can synthesize constraints by searching for source expressions in a dictionary or a terminology database. Dictionaries generally map one expression to many target ones and we or the model have to decide which of them to use. Terminology databases are usually unambiguous and the target translation serves as the constraint.
We experiment with terminology in~\cref{sec:term}.
 
 \paragraph{Lemmatization} Our methods use lemmatized\footnote{In Appendix B, we show that simple stemming heuristic performs at least as well as proper lemmatization in automated metrics described further.} constraints. For the random target subsequence method, we lemmatize the selected words. For the dictionary search method, we lemmatize both the dictionary and training data and we search for matching expression pairs using the lemmas. During the actual training, we use the original, non-lemmatized sentence with lemmatized constraints. This scenario is more similar to real-life use cases, since target word form which should be produced is not known beforehand. With constraint lemmatization, the above example would be:
 \jp{This one is interesting, but if we need space, we can convert it into plain text.}
 \begin{tabular}{p{7.4cm}}
    \\
        Input: Price increase is planned  mainly in larger municipalities. \texttt{<sep>}  \textbf{obec}  \texttt{<c>} \textbf{plánovat}\\ \\

 \end{tabular}
 \vspace{-0.5cm}
\section{Experiments}
\label{sec:experiments}
In this section, methods presented above are compared on various tasks and datasets. First, we use an oracle test set, which is created with previous knowledge of the reference.
We use it to assess the ability of the models to integrate the constraints themselves without additional noise caused by problems of the real world.
In the subsequent experiments, we present a more realistic scenario -- we use official terminology for EU-related expressions to translate parts of Europarl corpus.
Finally, we evaluate the approaches on translation of general, open-domain rare words using dictionary.


\subsection{Data}
\label{sec:data}
We train English-Czech NMT models for our experiments.
Czech has a high degree of inflection with seven cases and three genders for nouns and adjectives.
We train our models on CzEng 2.0~\citep{kocmi-2020-announcing} using all authentic parallel sentences (61M), as well as back-translated Czech monolingual sentences (51M).
Newstest-2019 \cite{barrault-etal-2019-findings} is used as a validation set and newstest-2020 \cite{barrault-etal-2020-findings} as a test set.
We break the text into subwords using SentencePiece~\cite{kudo-richardson-2018-sentencepiece} and lemmatize using UDPipe~\cite{straka-strakova-2017-tokenizing}. BLEU scores are computed using SacreBLEU \cite{post-2018-call}.\footnote{SacreBLEU signature: BLEU+case.mixed+lang.en-cs+numrefs.1+smooth.exp+test.wmt20+tok.13a+version.1.4.14}

For experiments mentioning dictionaries, we extracted pairs of terms from English and Czech Wiktionary\footnote{\url{www.wiktionary.org}} and a large commercial dictionary.
In  appendix \ref{app:wiki} we show that using Wiktionary also improves performance upon baseline, but the commercial dictionary offers better coverage of the expressions and thus provides better overall results. For this reason, all the experimets shown further are based on the commercial dictionary data.


We use the Czech government database for EU terminology\footnote{\url{sap.vlada.cz/dul/zavaznet.nsf/ca?OpenView}} to evaluate integration of domain-specific terminology through constraints.
We select all Czech terms and their translation to English, which corresponds to 14203 expressions per language.
Then, we search the Europarl\footnote{\url{www.statmt.org/europarl/}} corpus~\cite{koehn-2005-europarl} for sentence pairs containing English terms in the source side and lemmas of the Czech translation in a lemmatized version of the target side, ignoring trivial terms.
Keeping at most the first ten sentence pairs containing specific source term, the final dataset consists of 6585 examples, covering 1433 terms.
We remove these sentences from the training data, since Europarl is part of the CzEng corpus.

\subsubsection{Model}

We use MarianNMT~\cite{junczys-dowmunt-etal-2018-marian-fast} to train Transformer-base models with standard parameters~\cite{vaswani-2017-attention}.
Inspired by \citet{biblio:PoToTransformingmachine2020}, we alternate between authentic and backtranslated data every 25 million training sentences, while using exponential smoothing of the parameters.
Four NVIDIA V100 GPUs were used for the training and one training run (400-500k steps) takes approximately 40 hours with this configuration. A large portion of the computation time can be saved by finetuning an existing NMT model on the proposed dataset. By finetuning the baseline model we reached the same performance after 30-50k steps. However, all the results provided in this paper are obtained by training from scratch.
Since we integrate constraints in the target language into the source sequence, we share source and target vocabularies (and embeddings), consisting of 32000 subwords, to allow easier copying of the subwords from source to target sequence.

\begin{table*}[t]
\begin{center}
\begin{tabular}{lllcccc}
\textbf{Train const.} & \textbf{Train form}  &\textbf{Test form} & \textbf{BLEU} & \textbf{Cvg} & \textbf{BLEU$_L$} & \textbf{Cvg$_L$}\\
\toprule
baseline & - & - & 32.0 & 68.84 &	38.2 &78.14 \\


\hline


random & -  & - & 31.2 & 69.59&	37.1 &78.47  \\
random & surface & surface & 34.5 & \textbf{94.00} &	39.9 &\textbf{94.55}\\ 
random & surface & lemma & 27.1	& 61.31 &	36.8 &94.26 \\
random & lemma & lemma &  33.3 &	82.37 &39.7 &93.61 \\
\hline
dict & surface  & surface & 16.5&	57.34&	20.4	& 68.69  \\
dict & surface &  surface & \textbf{37.7} & 93.46 &42.2 &93.23  \\
dict & surface &  lemma  & 30.6& 64.11 &	39.6 & 91.55 \\
dict & lemma & lemma &  34.2 & 78.61 &	40.5 & 89.02  \\
\hline
dict, skip half & surface  & - & 31.7	& 68.88 &	38.2 & 78.06  \\
dict, skip half & surface &  surface & 36.9 & 91.37 &	\textbf{42.3} & 93.00  \\
dict, skip half & surface &  lemma  & 31.4	& 68.0&	40.0 &90.79 \\ 
dict, skip half & lemma & lemma & 33.1 & 75.36 & 39.3 & 85.30 \\
\hline


\end{tabular}
\caption{Results on newstest-2020 with oracle constraints. The first column shows the methods used for obtaining constraints for training. `random' means sampling random subsequences of target tokens, `dict' stands for terms matched by dictionary. In the `skip half' variant, a half of the training examples is presented with no constraint. For test sets, only constraints from the dictionary are used, still chosen so that the reference sentence contains the requested words. The second and third column indicate if the appended constraints are lemmatized or not, at training and test time, respectively.  
}
\label{tab:gold}
\end{center}
\end{table*}

\subsection{Oracle constraints}
\label{sec:oracle}

To assess the ability of a model to produce the provided constraints in the output, we use newstest-2020 test set with oracle constraints.
These constraints are obtained via dictionary search on the test set as described above, \textit{i.e.}, the constraints are terms from a English-Czech dictionary, where both source and target sides are present in the sentence pair.
Note that we know the reference beforehand, thus, this evaluation may not reflect improvement in translation in a real world setting.
We only use it to measure the ability of constraint integration.

We trained two sets of constrained models. The first one, baseline constrained models, use original target side forms of the constraint expressions. The second set consists of models trained using lemmatized forms of the constraints. Our goal with the lemmatized models was to harness the language modeling capacity of the model to generate a surface form of lemmatized constraint that agrees with the rest of the translation.

\cref{tab:gold} presents the results. We used two forms of the test set constraints -- original reference forms and lemmatized constraints (column \textit{Test form}). The lemmatized constraints are closer to real world scenario, where we do not know the output form of the constraint expression beforehand.

As a sanity check, we compute standard BLEU and BLEU calculated on lemmatized hypothesis against lemmatized reference ($\textit{BLEU}_L)$.
More importantly, we assess target constraint coverage (\textit{Cvg} and $\textit{Cvg}_L$) on original and lemmatized test set by comparing the constraints in the output with the reference.   Note that in theory, \textit{Cvg} value should always be lower or equal to $\textit{Cvg}_L$, since surface form coverage is equal to lemma coverage minus proportion of incorrectly generated surface forms.  This is not always the case, since the lemmatizer takes the sentence context into consideration and lemmatized versions of stand-alone terms in the terminology database may not match lemmatized versions of the same terms inside a reference sentence. This causes a slight underestimation of $\textit{Cvg}_L$.

The \textit{Cvg} and $\textit{Cvg}_L$ columns document that both methods of constraint synthesis for training (random target subsequences and dictionary terms)
lead to models capable of producing more than 93\% of the constraints when constraints are not lemmatized.
Surface coverage of surface form trained models drops to 61--68\% when using lemmatized form of the test set constraints, but lemma coverage is only slightly lower -- this is expected, as these models are trained to reproduce exact form of the given constraints.

The results of models trained on lemmatized constraints with lemmatized test constraints show that the surface form coverage increases compared to surface form trained models with lemmatized test constraints (rows \textit{lemma}/\textit{lemma} vs. \textit{surface}/\textit{lemma}).  While the  coverage is lower than when using surface form test set for the surface form model, we show in \cref{sec:analysis} that this is mainly an artifact of reference-based evaluation and that the model inflects the constraints correctly.

The model trained with constraints based on dictionary reaches the best performance on the oracle constraint test set, for which the constraints are generated in the same way.
However, when constraints are not supplied, BLEU and coverage drops sharply (the row \textit{dict/surface/-}).
This may be caused by the fact that sentences containing expressions present in the dictionary are almost always accompanied by the constraint during the training.
Therefore, the model is not presented with many examples where the translation appears without the corresponding constraint and generates constraint expression with much lower probability when this happens during the test time.
We experimented with skipping half of the sentences during the constraint generation, leaving them without any constraints (``skip half'' in  the table).
As shown in \cref{tab:gold}, this largely reduces the problem -- without any test time constraints, the model reaches baseline results (the row \textit{dict, skip half/surface/-}).
However, when the constraints are supplied, the coverage is slightly lower than for a model trained with constraints for all the sentences (e.g. 91.4\% instead of 93.5\% for surface form models).
Fine-tuning the ratio or choosing the sentences to leave without the constraints dynamically during the training might help to solve this problem.

\begin{table}[t]
\small
\centering
\begin{tabular}{llcc}
\textbf{Train}     & \textbf{Test}  & \textbf{BLEU} & \textbf{Cvg} \\
\toprule

Baseline & No constraints&37.9&75.02 \\
\hline

\multirow{4}{*}{All}               & No constraints        & 19.1 & 61.40     \\
& Terms &  37.3& 91.73 \\
                       & Dict    &43.3         &    84.14\\
& Terms + Dict  & 44.0& 93.75\\
\hline
\multirow{4}{*}{Skip half}         & No constraints        & 38.2 & 75.32   \\

                       & Terms            & 38.4  & 90.52   \\
                                              & Dict    &43.5         &    83.49\\

                       & Terms + Dict    &43.1         &    91.22\\
                       \hline

\end{tabular}
\caption{Performance of models trained using \textbf{surface forms} of \textbf{dictionary constraints} on the \textbf{same} Europarl test set split. \textit{Train} column documents whether all of the training sentences were accompanied by constraints, or we left 50\% of them without constraints (\textit{Skip half}). \textit{Term} constraints come from a terminology database, \textit{Dict} constraints are expressions from a general dictionary. Note that for applying \textit{Dict} constraints at test time, we used test reference for dictionary target term disambiguation, which makes this combined approach not feasible in realistic conditions. All test set constraints are used in reference surface forms.
}
\vspace{-0.5cm}

\label{tab:europarl_dict}
\end{table}

\subsection{Terminology Integration}
\label{subsec:europarl}

\label{sec:term}
Since the studied methods proved to work well with oracle surface form of constraints, we moved to a realistic use-case with the Europarl test set described in Section~\ref{sec:data}.  
We split the test set into two parts:

\begin{itemize}
    \item    \textbf{\textit{same}}   contains examples where the form of the constraint  in the reference is the same as in the terminology database (and as provided to the baseline constrained model),
     \item \textbf{\textit{diff}}  
     contains examples where the form of the constraint in the target sentence is different from the database form.
\end{itemize}
The target lemmas of the constraint should match in both cases.

This split allows us to better assess the translation in inflected languages, since the problems we focus on are more pronounced in the \textbf{diff} test set.
\cref{tab:europarl_dict} shows that the model trained with dictionary constraints underperforms in terms of BLEU when only the constraints from terminology database are supplied (BLEU of 19.1).
This is caused by the issue described earlier -- during the training, the model does not encounter the words which are present in the dictionary enough times without the constraint. When the dictionary constraints are used alongside the terminology database constraints (rows denoted by ``Terms + Dict''), the BLEU score increases.
This approach requires either prior knowledge of the reference,  or a mechanism for the target dictionary term disambiguation.
To mitigate this issue, we skip half of the sentences when generating the constraints, \textit{i.e.}, half of the training corpus is seen without any constraints.
This alleviates the problem to a large extent, see the ``Skip half'' results.

\begin{table}[t]
\centering
\small
\setlength{\tabcolsep}{5pt}
\begin{tabular}{llccc}
\textbf{Train c. }        & \textbf{Test c. }& \textbf{BLEU} & \textbf{Cvg} & \textbf{Cvg$_L$}\\
\toprule
-               & -        & 38.2 & 69.90  & 84.37  \\
\hline

\multirow{4}{*}{SF}        & -        &  38.8 & 70.27 & 85.0 \\
&canon. & 36.6 &	44.0  & 96.56\\
                      &  Ref SF &    40.6 &	96.97 & 95.08 \\
                       & lemma            & 35.1	&30.88 & 96.74   \\
                       \hline

\multirow{4}{*}{Lemma}              & -        & 38.6 & 69.87 & 84.05   \\
                        &canon. &  38.9 &77.1 &	95.44  \\
                       & Ref SF            &  39.1	& 81.44 & 94.15   \\
                       & lemma            &   38.9 & 77.22   & 95.55  \\
                       \hline

\end{tabular}
\caption{Results on \textbf{whole} Europarl test set.  None of the BLEU scores for constrained models (except \textit{Ref SF}) is significantly better than the best unconstrained score. }
\vspace{-0.35cm}
\label{tab:europarl_whole}
\end{table}


We present the results on the whole test set in Table~\ref{tab:europarl_whole}.
The first and second columns show word form of the constraints during the training and test time, respectively.
\textit{Canon.}  constraint is in its canonical, original form from the the terminology database.
\textit{Ref SF} rows show results with constraints in the same form as in the reference translation (this requires prior knowledge of the reference).

First, let us focus on results of models trained with surface form constraints.
Three trends in the results hint that generating the correct constraint form is challenging for the model, if the correct form is different from the one supplied in the input.
First, the difference between surface form and lemma coverage (44\% vs 96.6\%) shows the model generates the correct constraint words, but in a form not matching the reference. 
Second, the difference is more pronounced in the \textbf{\textit{diff}} split (\cref{tab:europarl_diff}), while in the \textbf{\textit{same}} split (\cref{tab:europarl_same}), surface form coverage is almost the same as the lemma coverage.
This is because in the \textbf{\textit{same}} split, target constraints are already in the canonical form, same as in the terminology database, so there is no need for further inflection.
Third, using constraints in the same surface form as in the reference (\textit{Ref SF}) improves the observed coverage compared to using the canonical form from the terminology database (e.g., 97\% vs 44\% on the whole test set, see \cref{tab:europarl_whole}). 
This ``oracle'' setting, using the reference to determine the correct surface form, shows the upper limits of the constraint integration approach, if the inflection issue is solved optimally.

\begin{table}[t]
\setlength{\tabcolsep}{5pt}
\centering
\small
\begin{tabular}{llccc}
\textbf{Train c. }        & \textbf{Test c. }& \textbf{BLEU} & \textbf{Cvg} & \textbf{Cvg$_{L}$} \\
\toprule
-               & -        & 38.3 & 67.1& 84.12    \\
\hline

\multirow{4}{*}{SF}         & -        & 38.8 & 67.14  & 84.68 \\
                       & canon.                     & 35.0 & 15.20 & 96.20     \\
                       & Ref SF            & 40.8& 96.32 & 93.92    \\
                       & lemma            & 34.3 & 15.38 &  96.41   \\
                       \hline

 \multirow{4}{*}{Lemma}  & -        & 38.7   & 66.61  & 83.42  \\
                       &    canon.                  & 38.9 & 72.31 & 94.76    \\
                       & Ref SF            & 39.2 & 79.16  & 92.78   \\
                       & lemma            & 39.0 & 72.62   & 94.88 \\

\end{tabular}
\caption{Results on \textbf{diff} Europarl test set split, where we only consider cases where the constraint is provided in different form than in the reference, i.e. reference contains different form than the canonical one present in the terminology database. None of the BLEU scores for constrained models (except \textit{Ref SF}) is significantly better than the best unconstrained score.}
\label{tab:europarl_diff}
\end{table}

\begin{table}[t]
\setlength{\tabcolsep}{5pt}
\centering
\small
\begin{tabular}{llccc}
\textbf{Train c. }        & \textbf{Test c. }& \textbf{BLEU} & \textbf{Cvg} & \textbf{Cvg$_L$}\\
\toprule
-               & -        & 37.9 & 75.02  &  84.72\\
\hline

\multirow{3}{*}{SF}        & -        & 38.8 & 75.94&	85.50    \\
                       & canon.        &    \textbf{39.9} &	97.69	& 97.03   \\
                       & lemma            & 36.6 & 59.56 & 97.38   \\
                       \hline
        \multirow{4}{*}{Lemma}  & -       & 38.4& 75.89 & 85.15  \\
                       & canon.            &38.8 & 85.81    & 96.55\\
                       & lemma            &38.8 & 85.58 & 96.55   \\

\end{tabular}
\caption{Results on \textbf{same} Europarl test set split.
In this subset, the constraints from terminology database are already in the same form as in reference, \textit{i.e.} \textit{canon.} is the same as \textit{Ref SF}. BLEU score that is significantly better than the best BLEU without constraints is in bold (bootstrap resampling, $p\leq0.05$).}
\label{tab:europarl_same}
\vspace{-0.4cm}

\end{table}


As stated earlier, we trained the models again using lemmatized versions of the constraints.
When we supply lemmatized constraints to these models during the test time, the coverage rises from 44\% (surface form trained model with canonical constraint forms) to 77\%, but this is still far from the oracle 97\%.
This suggests that a large room for improvement exists, but as we show in Section~\ref{sec:analysis}, most of these discrepancies are caused by reference-based evaluation and are not real errors.
In majority (92\%) of the cases marked as not covered when using lemmatized model, the form of the constraint is different from the reference, but correct given the context, as the model translates the sentences differently (but correctly).

\subsection{Comparison with constrained decoding}

Our work is based on training the NMT model to include provided constraints in the output translation. Another popular way of constraint integration is modifying the decoding process. We hypothesize that this approach will not be useful in our scenario, since the constraints are enforced in their surface forms, which is the issue we are trying to solve. To verify this, we evaluated lexically constrained decoding by \citet{hu-etal-2019-improved} as implemented in fairseq~\cite{ott-etal-2019-fairseq} on the Europarl test sets described in Section \ref{subsec:europarl}. 

\begin{table}[ht]
\begin{center}
\small
\setlength\tabcolsep{4.5pt}
\begin{tabular}{llccccc}
\textbf{Split} & \textbf{Con.}   & \textbf{BLEU} & \textbf{Cvg} &\textbf{BLEU$_L$} & \textbf{Cvg$_L$ } &\textbf{Pos $\rho$}\\

\toprule

Same & no &   36.4 & 69.3  &  42.8 & 79.7  & 0.95 	\\
Same & yes & 35.7  & 97.1 & 41.5 & 97.3	&  0.83\\

Diff & no & 36.4  & 63.1	  & 43.1 & 81.3 & 0.95 \\
Diff & yes &   30.6 & 26.0   & 39.3  & 94.7 & 0.80	\\

Whole & no &  36.4  & 65.2  & 43.0 & 80.8 & 0.95	\\
Whole & yes & 32.3 & 50.7 & 40.0 & 95.6 & 0.81	\\

\end{tabular}
\end{center}
\caption{Lexically constrained decoding}
\label{tab:fairseq}
\end{table}
The results in Table~\ref{tab:fairseq} show that while the constrained decoding indeed produces the target constraints in the output, they stay in the same form as in the terminology database. This is shown by the low surface form constraint coverage (column \textit{Cvg}) for the \textbf{\textit{diff}} and \textbf{\textit{whole} }dataset splits, while for the \textbf{\textit{same} }split, where the constraints are in the same form in the translation as in the terminology database, the coverage is high. On lemma level (\textit{Cvg}$_L$), coverage on all splits remains high, again showing that the system produces exactly the surface form provided, instead of correct target sentence form. Note that the results are not directly comparable with the results in previous subsection, since here we use only a part of the training data (first 25M sentence pairs from parallel part of CzEng) for the preliminary experiments.

We also observed that the Pearson correlation of constraint placement in respect to reference translation  (see Appendix \ref{app:place} for details) is lower (0.81) when using constrained decoding  than when using the training approach as in the main experiments (0.94).

\subsection{Semi-parametric rare words translation}
\begin{table}[t]
\small
\begin{center}
    \begin{tabular}{lccc}
    \textbf{Constraint src } & \textbf{BLEU} & \textbf{\% as ref }& \textbf{\% correct} \\
    \hline 
    No constraint  & 21.6 & 35.4& 64.6 \\
    Reference term & 23.1 & 91.7 & 91.7 \\
    Random  term &  22.6 & 54.2 & 83.3\\
    \end{tabular}
    \caption{Translation of sentences containing rare words. For source expressions with multiple possible translations according to the dictionary, we compare choosing a translation variant randomly (\textit{Random term}) against choosing the same translation variant as in the reference. All  constraints are lemmatized. Column \textit{\% as ref} shows the percentage of examples with the constraint translated with the same term as in the reference. Column \textit{\% correct} shows human evaluation of rare word translation. 
    }
    \label{tab:rare}
    \vspace{-0.5cm}

\end{center}
\end{table}

We define rare words as terms from a dictionary that occur in the source side of the training corpus at most 50 times.
We create a subset of our general dictionary by only using expression pairs with rare words on source side.
We search WMT 2007-2020 English-Czech news test sets  \cite{barrault-etal-2020-findings} for sentence pairs containing term pairs from this rare word dictionary, resulting in 48 examples.
A dictionary generally provides 1-to-many mappings of source terms to a target language, so the correct target expression needs to be disambiguated.
Table~\ref{tab:rare} presents results with no constraints, with constraints where the lemmatized target constraint is chosen based on the lemmatized reference, and with constraints where the target expression is chosen randomly from all the possible translations. We used a model trained on lemmatized random target token subsequences for the translation.
On average, each rare word in the test set has 3.3 possible dictionary translations.
Aside from BLEU score, we show the percentage of rare words translated correctly, meaning that either they are the same expression as in the reference, or that they are synonymous expressions that are correct in the given context. This is different from the terminology use case, since we do not strictly enforce single possible translation.
The results show that even with the random choice of the dictionary constraint translation, our model improves the translation of rare words.

\section{Manual analysis}
\label{sec:analysis}
\begin{table*}[h]
    \centering
    \small
    \begin{tabular}{p{0.24\linewidth}|p{0.05\linewidth}|p{0.05\linewidth}|p{0.44\linewidth}|l}
    \textbf{Source} & \textbf{Canon} & \textbf{Ref} & \textbf{Translation} & \textbf{Error} \\\hline
    \multirow{3}{\linewidth}{They are seeking to weaken the Commission's proposal to benefit the industry.} & \multirow{3}{*}{návrh}& \multirow{3}{*} {návrhu} &  Snaží se oslabit \textbf{návrh} Komise ve prospěch průmyslu. & CCC \\
    &&& Snaží se oslabit \textbf{návrh} Komise na prospěch průmyslu. & CIC \\
    &&& Snaží se oslabit \textbf{návrhu} Komise ve prospěch průmyslu. & Inflection \\\hline
    \end{tabular}
    \caption{Example of three error types given canonical and reference target form constraints.}
    \label{tab:err_examples}
\end{table*}

In this section, we analyse examples marked as errors by automatic evaluation. In \cref{app:place}, we analyse the position of constraints in translation outputs, showing that they are placed correctly. In  \cref{app:lease}, we look closely at the constrained translation of an out-of-domain document. 

\begin{table}[t]
\begin{center}
    
\small
\begin{tabular}{llll}
\textbf{Error type}        & \textbf{B } & \textbf{SF}& \textbf{L} \\
\toprule
Incorrect inflection & 2 & 46 & 0\\
Correct in correct context      & 65 &44 & 92        \\
Correct in incorrect context   & 0   &     3 & 2      \\
Different correct word choice & 28 &2& 4 \\ 
Different incorrect word choice & 0 & 0& 1 \\
Invalid translation&5 &5 & 1
\end{tabular}
\caption{Analysis of 100 outputs marked as errors by the automatic evaluation, which means that either they do not contain the constraint or they contain it in a different surface form compared to the reference. We analysed three models -- baseline (\textbf{B}), a model trained with surface form constraints using canonical forms of the constraints at test time (\textbf{SF}), and a model trained with lemmatized constraints using lemmatized terminology entries at test time (\textbf{L}).
}
\label{tab:europarl_errors}
\end{center}
\vspace{-0.55cm}
\end{table}

\subsection{Error analysis}

We manually analysed outputs marked as not having the desired constraint in the reference surface form by the automatic coverage evaluation introduced in the previous section.
Table~\ref{tab:europarl_errors} presents the results.
We compare three models.
First, the baseline without any constraints (column \textbf{B}).
Second, the best model trained with non-lemmatized constraints (\textbf{SF}), and, finally, the best model trained on lemmatized constraints (column \textbf{L}).
The baseline model outputs have constraint surface form coverage of 69.9\% on the whole Europarl test set, which results in 1982 out of 6585 examples being marked as different from the reference by the automatic evaluation.
The SF model reached 44\% coverage (4346 differences).
The lemmatized model agreed with the reference in 77.1\% (1508 differences).
For each model, we randomly sample 100 supposedly erroneous translations to be analysed.

The first row of \cref{tab:europarl_errors} shows the number of examples with constraints incorrectly inflected in the context of the generated output.
Rows 2 and 3 show cases where the constraint form agrees with rest of the translation:
\textit{Correct in correct context (CCC)}  indicates that the target sentence is a valid translation,
whereas \textit{Correct in incorrect context (CIC)}  indicates that the constraint was inflected correctly given its context but as a whole, the translation is wrong.
Thus, \textit{CCC} cases are not in fact errors, but were wrongly classified as such by the automatic evaluation, based on a direct comparison with the reference. 
The cases where the model ignores the constraint and generates a different word are in the categories \textit{Different correct/incorrect word choice} (fourth and fifth rows), based on whether the generated word is a plausible translation of the source constraint.
Examples where the translation generally goes wrong and the issue does not fit into the previous categories are under \textit{Invalid translation}.

Our analysis shows that for the lemmatized model (\textbf{L}), the vast majority of the examples classified as errors are actually correctly translated and contain the requested constraint in the correct surface form. The presumed error is an artifact of the reference-based evaluation. Only 8\% of these examples are real errors, compared to 66\% for the surface form model. 

In \cref{tab:err_examples}, we show three examples of errors found by the automatic evaluation.
Given the canonical and reference source form of a constraint (návrh and návrhu, respectively, meaning ``proposal''), some errors may arise in the translation.
In the first row, although different from the reference source form, the constraint is correctly inflected given the context generated and in a correct translation, which configures a ``correct in correct context'' error (CCC).
Similarly, in the second row, the same constraint with the same source form is correctly inflected given the context but in a wrong translation, which describes a ``correct in incorrect context'' (CIC) error.
Finally, the third translation has a wrong inflection given the context generated (Inflection error).

\section{Conclusion}
\label{sec:conclusion}

We described the problem of word inflection in lexically constrained machine translation.
Our solution capitalizes on the ability of NMT models to generate correct word forms in the output translation.
We train a Transformer model using lemmatized constraints supplied alongside the input sentences, and correct surface forms of the constraints in the reference.
This training leads to a model producing the constraints in the output with high coverage, correct placement, and in a correct surface form.

We compare several methods of obtaining constraints and integrating them into the input.
In the realistic use case of terminology integration, we evaluated our methods and show that without lemmatizing the training constraints, the chosen approach of integrating constraints into NMT does not work well for Czech. 
We effectively solve the issue of inflection errors by lemmatizing constraints, taking advantage of the Transformer's language modelling capacity with no additional inference costs.
This has been proven by both automatic and manual evaluation.
We show our method is also effective in translating general domain rare words using a bilingual dictionary and we plan future work in solving the problem of choosing correct translation term from number of variants.

\section*{Acknowledgements}
Our work is supported by the Bergamot project (European Union’s Horizon 2020 research and innovation programme under grant agreement No 825303) aiming for fast and private user-side browser translation, GA ČR NEUREM3 grant (Neural Representations in Multi-modal and Multi-lingual Modelling, 19-26934X (RIV: GX19-26934X)) and by SVV 260 453 grant.
We also want to thank Michal Novák for his useful feedback and discussions.

\newpage

\bibliography{custom}
\bibliographystyle{acl_natbib}
\clearpage

\appendix 
\section{Further analysis}
\label{app:a}

\subsection{Constraint placement}
\label{app:place}

Increased BLEU and constraint coverage show that the evaluated methods are able to generate correct constraint string in the output.
However, these metrics do not tell much about placement of constraints.
If all the constraints are appended at the end of the output, we would get perfect coverage and, in some cases, possible increase in BLEU score -- but this is  not a desired behavior of the system.

To evaluate the correctness of constraint placement, we record starting indices of each satisfied constraint in both MT output and reference, and we compute Pearson's correlation between these two variables.
As a sanity check of the correlation measure, we also modify the output of the constrained system and move the constraints it correctly produced to random positions. Both BLEU and the Pearson correlation drop considerably, see the line marked with ``*'' in \cref{tab:placement}.

The second row shows the case of supplying constraints as a suffix for the baseline model, which was not trained to utilize them.
Coverage of the constraints has increased -- but, as expected, the model only generates some of the constraints at the end of the translation.
Lower correlation with the reference placement shows that the placement is incorrect.
In the fourth row, we randomly change positions of the constraints as described above.
Again, the correlation decreases.
These experiments indicate that the evaluated systems can generate constraints at correct positions in the output. 

\subsection{Lease agreement case study}
\label{app:lease}

Our method proved to work well on the Europarl terminology test set. 
Since Europarl is included in the training data (only the actual test sentences are filtered out), we used an out-of-domain test document to assess the results using unknown terminology.
For this purpose, we used a sublease agreement translated from Czech into English, which is included in WMT20 Markables test suite\footnote{\url{https://github.com/ELITR/wmt20-elitr-testsuite/}} \cite{zouhar-etal-2020-wmt20}.
There are minor translation errors in the reference of the test set version used at WMT20, which we fixed.\footnote{We will provide the link to the fixed test set in the camera-ready version.}\jp{We need to modify this footnote to add the location of the fixed test set.}
The difficulty of translating this agreement accurately lies in the translation of some of the legal terms, e.g. \textit{tenant}, \textit{lessee}, \textit{lease} or \textit{sublease}. These terms are often used interchangeably in common language.
In this case, tenant (\textit{nájemník} in Czech) is a person who has an apartment in a lease from the owner and lessee (\textit{podnájemník}) is a person that is using the apartment based on the agreement with the tenant. 

\begin{table}[t]
\begin{center}
\begin{tabular}{llccc}
\textbf{Model} & \textbf{Constr}. & \textbf{BLEU} & \textbf{Cvg} & \textbf{Pos} $\rho$ \\
\hline 
Baseline & - &  30.9 & 70.70 & 0.9362\\
Baseline & suffix &  27.6 &76.93 & 0.8407\\
Suffix & suffix &  35.3 &95.05 & 0.9382 \\
Suffix * & suffix & 16.8  &95.05 & 0.3486 \\
\end{tabular}
\caption{Correlation between start character indices of the satisfied constraints in system's output and reference. The table shows that the evaluated methods place constraints at the correct positions in the output. When we move the constraints (marked with an asterisk), the correlation between the positions drops.
\mn{I moved this table here in order to be placed in the right column of the first page of Appendix.}
}
\label{tab:placement}
\end{center}
\end{table}

\begin{table*}[ht]
\small
\begin{tabular}{lp{0.79\linewidth}}
\textbf{Model}      & \textbf{Translation}         \\                                                                                                \toprule
                                                                                                       
Source              & In Art. III of the $\textbf{Sublease agreement}_{1}$, entitled “ $\textbf{Term of the Lease}_{2}$ ,” the $\textbf{tenant}_{3}$, and the $\textbf{lessee}_4$ agreed that the $\textbf{apartment in question}_5$ would be rented to the $\textbf{lessee}_6$ for a fixed period from 13th May 2016 to 31st December 2018. \\\\\hline
Google Translate    & V čl. III $\textcolor{darkgreen}{\textbf{smlouvy o podnájmu}_1}$ s názvem „$\textcolor{red}{\textbf{Doba nájmu“}_2}$ se $\textcolor{darkgreen}{\textbf{nájemce}_3}$ a $\textcolor{red}{\textbf{nájemce}_{4}}$ dohodli, že $\textcolor{darkgreen}{\textbf{předmětný byt}_5}$ bude $\textcolor{red}{\textbf{nájemci}_{6}}$ pronajat na dobu určitou od 13. května 2016 do 31. prosince 2018.                                                    \\\\
Lingea Translator   & V čl. III $\textcolor{orange}{\textbf{podnájemní smlouvy}_1}$, nadepsané „$\textcolor{red}{\textbf{Lhůta nájmu}_2}$ ,„ se $\textcolor{darkgreen}{\textbf{nájemce}_3}$ a $\textcolor{red}{\textbf{nájemce}_{4}}$ dohodli, že $\textcolor{orange}{\textbf{dotčený byt}_5}$ bude $\textcolor{red}{\textbf{nájemci}_6}$ pronajat na dobu určitou od 13. května 2016 do 31. prosince 2018.                                               \\\\
CUBBITT             & V umění. III $\textcolor{orange}{\textbf{podnájemní smlouvy}_1}$ nazvané „ $\textcolor{red}{\textbf{Podmínky pronájmu}_2}$ “ se $\textcolor{darkgreen}{\textbf{nájemce}_3}$ a $\textcolor{red}{\textbf{nájemce}_4}$ dohodli, že $\textcolor{darkgreen}{\textbf{předmětný byt}_5}$ bude $\textcolor{red}{\textbf{nájemci}_6}$ pronajímán na dobu určitou od 13. května 2016 do 31. prosince 2018                                        \\\\\hline
Suffix surface form & V čl. III $\textcolor{red}{\textbf{podnájemní smlouva o podnájmu}_1}$, nadepsaném „ $\textcolor{orange}{\textbf{Lhůta nájmu}_2}$“, se $\textcolor{darkgreen}{\textbf{nájemkyně}_3}$ a $\textcolor{darkgreen}{\textbf{podnájemkyně}_4}$ dohodly, že $\textcolor{darkgreen}{\textbf{předmětný byt}_5}$ bude $\textcolor{red}{\textbf{nájemci}_6}$ pronajat na dobu určitou od 13. května 2016 do 31. prosince 2018.                            \\\\
Suffix lemmatized   & V článku III $\textcolor{darkgreen}{\textbf{smlouvy o podnájmu}_1}$, nazvaném „$\textcolor{darkgreen}{\textbf{doba podnájmu}_2}$ ,“ se $\textcolor{darkgreen}{\textbf{nájemkyně}_3}$ a $\textcolor{darkgreen}{\textbf{podnájemkyně dohodli}_4}$, že $\textcolor{darkgreen}{\textbf{předmětný byt}_5}$ bude $\textcolor{red}{\textbf{nájemci}_6}$ pronajat na dobu určitou od 13. května 2016 do 31. prosince 2018.                                  \\\\
Factored SF         &   V čl. III $\textcolor{darkgreen}{\textbf{smlouvy o podnájmu bytu}_1}$, nadepsaný „$\textcolor{red}{\textbf{podnájmu}_2}$ ,“ 
$\textcolor{darkgreen}{\textbf{nájemkyně}_3}$ a $\textcolor{darkgreen}{\textbf{podnájemkyně}_4}$souhlasily s tím, že $\textcolor{darkgreen}{\textbf{předmětný byt}_5}$ 
bude pronajat $\textcolor{darkgreen}{\textbf{podnájemkyni}_6}$ na dobu určitou od 13. května 2016 do 31. prosince 2018.
                                                                    \\\\\hline
Ref                 & V čl. III $\textcolor{darkgreen}{\textbf{Smlouvy o podnájmu}_1}$ bytu, nazvaném „$\textcolor{darkgreen}{\textbf{Doba podnájmu}_2}$“, se $\textcolor{darkgreen}{\textbf{nájemkyně}_3}$ a $\textcolor{darkgreen}{\textbf{podnájemkyně}_4}$ dohodly, že $\textcolor{darkgreen}{\textbf{předmětný byt}_5}$ bude $\textcolor{darkgreen}{\textbf{podnájemkyni}_6}$ přenechán k užívání na dobu určitou od 13. 5. 2016 do 31. 12. 2018.                           
\end{tabular}
\caption{Translations of one of the difficult sentences from WMT20 ELITR test set.}
\label{tab:lease}
\end{table*}

 We manually created a database of 11 legal terms and their translations used in the document. Note that we know that the sublease agreement is between two women, so we used feminine forms of the translations for lessee and tenant.
 Table~\ref{tab:lease} compares translations produced by our systems against existing approaches on one problematic sentence. We used following term pairs as our constraints for this sentence:
 
 \begin{tabular}{p{0.43\linewidth} p{0.44\linewidth}}
 \\
 \textbf{Source} & \textbf{Target} \\
 \toprule
Term of the Lease&Doba podnájmu \\
lessee&podnájemkyně \\
tenant&nájemkyně \\
apartment in question&předmětný byt \\
Sublease agreement&Smlouva o podnájmu bytu \\\\
 \end{tabular}
 
 Our three systems are: (1) the model based on suffixed surface form constraints, (2) the same model using lemmatized constraints, and (3) the two-factored model using surface form factors as described in \cref{app:factors}.
 They are compared with the outputs of CUBBITT\footnote{\url{https://lindat.mff.cuni.cz/services/translation/}}, the state-of-the-art English-Czech system  by~\citet{biblio:PoToTransformingmachine2020}, and two commercial translation engines (Google Translate\footnote{\url{https://translate.google.com/}} and Lingea Translator\footnote{\url{https://translator.lingea.com/}}).
 Constraint terms typeset in green are translated correctly according to the terminology, orange terms are very similar in meaning to the terminology database translation, and red ones are clear translation errors.
 We note that especially the word \textit{podnájemkyně} (\textit{lessee} in feminine form) poses some difficulties for the model to produce, since it does not appear in the training data.
 Its masculine forms, \textit{podnájemce} or \textit{podnájemník} appear 182 times in different inflections.


 Another difficulty is added by the fact that the word \textit{lessee} appears two times in the sentence. All of the systems produce the correct constrained translation at most for the first occurrence, with exception of factored model, which is supplied explicit alignment between source and target part of the constraints.
 We hypothesize that other models consider the constraint as covered after it is generated for the first time. 
 
 Overall, the constrained models provide more accurate translations compared to the unconstrained SOTA models, effectively integrating the constraints even in a difficult out-of-domain example.

\begin{table*}[ht]
\begin{center}
\begin{tabular}{lllcccc}
\textbf{Train const.} & \textbf{Integration}   & \textbf{BLEU} & \textbf{Cvg} &\textbf{BLEU$_L$} & \textbf{Cvg$_L$ } & \textbf{Pos $\rho$}\\
\toprule
baseline & - &   30.9 & 70.51 & 37.0 & 77.46 & 0.9322\\

random & prefix & 34.7 & \textbf{96.15} &39.4 &\textbf{95.51} & 0.8468\\
random & suffix & \textbf{34.9} & 93.02& 40.0 & 92.99 & 0.9336 \\

random, shift & suffix+shift & \textbf{34.9} & 93.12 & \textbf{40.1} & 93.25 &\textbf{0.9349}\\

\hline

\end{tabular}
\caption{Comparison of integrating the constraints as a prefix, suffix and suffix with positional embedding shifting. Note the results are not directly comparable to main paper results, as the train and test set preprocessing is different.
\mn{I commented out the rows random/suffix+shift a random,shift/suffix. They bring no valuable information which is confirmed by the fact that you do not comment on them.}
}
\label{tab:prefix}
\end{center}
\vspace{-0.6cm}
\end{table*}

\begin{table*}[h]
    \centering
    \small
    \begin{tabular}{|c|c|c|c|c|c|c|c|c|c|c|c|}
      \hline
        Words & Price & increase & is & \textit{planned} & \textbf{plánováno} &  mainly & in &  larger & \textit{municipalities}  & \textbf{obcích} &  . \\  
      \hline
        Factor  & 0 & 0 & 0 & SRC & TGT & 0 & 0 & 0 & SRC & TGT & 0\\\hline
    \end{tabular}
          
    \caption{Example of the constrained translation process using factors.}
    \label{tab:factors}
\end{table*}

\begin{table}[ht]
\centering
\small
\setlength{\tabcolsep}{5pt}
\begin{tabular}{llccc}
\textbf{Train c. }        & \textbf{Test c. }& \textbf{BLEU} & \textbf{Cvg} & \textbf{Cvg$_L$}\\
\toprule
-               & -        & 38.2 & 69.90  & 84.37  \\
\hline

\multirow{4}{*}{SF}        & -        &  38.8 & 70.27 & 85.0 \\
&canon. & 36.6 &	44.0  & 96.56\\
                      &  Ref SF &    40.6 &	96.97 & 95.08 \\
                       & lemma            & 35.1	&30.88 & 96.74   \\
                       \hline

\multirow{4}{*}{Lemma}              & -        & 38.6 & 69.87 & 84.05   \\
                        &canon. & 38.9&77.1 &	95.44  \\
                       & Ref SF            &  39.1	& 81.44 & 94.15   \\
                       & lemma            &  38.9 & 77.22   & 95.55  \\
                       \hline
            \multirow{4}{*}{Mixed}  
                        &  -        & 37.7 &	69.37 & 83.51 \\
                        &canon. & 37.5	& 69.68  & 95.08 \\
                       & Ref SF           & 39 &91.65 & 94.72    \\
                       & lemma            & 38	& 76.57 & 95.25 \\

\end{tabular}
\caption{Performance of the model mixing half of the training examples with surface form constraints and half of them lemmatized on the \textbf{whole} Europarl test set.
Compared with the models, where either lemmatization was never applied on constraints during training (SF), or it was applied on all data examples (Lemma).}

\label{tab:europarl_whole_mixed}
\end{table}

\label{app:preexp}

\section{Other related experiments}
\label{app:b}
\mn{I've never used Appendices so intensively, but I guess it is a good practice to refer to each of them from the main text. But referring too much would suggest that you were not able to squeeze the important information to 8 pages.}

We present experiments that influenced our architectural choices in the paper, but are not discussed in the main text. Note that the results are not directly comparable, since a slightly different preprocessing was used.

\subsection{Constraints as prefix or suffix}

In Table \ref{tab:prefix}, we compare passing the constraints as a prefix of the source sentence, as a suffix and as a suffix with all positional embeddings of the constraint part starting with 1024.
Using prefix resulted in the best coverage, but, as visible in column \textit{Pos $\rho$}, correlation of constraint positions is lower than for other models.
\mn{what is Pos? Is it Pearson's corrcoef as in the previous table. If so, use the same formatting of both the header and the values in the tables, including the following one.}
Upon manual inspection, we saw that the constraints were in some cases generated also as a prefix of the target sentence.
For the main experiments, we decided to use suffix integration with positional embedding shifting, since it provided slightly better coverage than the basic suffix variant.

\begin{table}[h]
\begin{center}
\small
\begin{tabular}{llcc}
\textbf{Training dict.} & \textbf{Test dict.}   & \textbf{BLEU} & \textbf{Cvg} \\

\toprule
- & Wiki &29.2  &79.5 \\
- & Large &29.2 &  69.2\\
Wiki & Wiki & 30.1 & 93.7  \\
Wiki & Large & 29.6 & 81.8 \\
Large & Wiki & 24.6 & 91.7 \\
Large & Large  &34.3 & 94.3 \\

\end{tabular}
\end{center}
\caption{Comparison between using large, commercial dictionary (\textit{Large}) as opposed to Wiktionary (\textit{Wiki}) to obtain both training and test constrains. The results are computed on the oracle newstest-2020 test set, see \ref{sec:oracle} for details. } 
\label{tab:wiki}

\end{table}

\begin{table*}[h]
\begin{center}
\begin{tabular}{lllcccc}
\textbf{Train const.} & \textbf{Integration}  &\textbf{ Const. form} & \textbf{BLEU} & \textbf{Cvg} & $\textbf{BLEU}_L$ & $\textbf{Cvge}_L$\\
\toprule
baseline & - & - &  30.9 & 70.70 & 37.1 & 77.73 \\
random & suffix & surface & 35.3 &95.05 & 40.4 & 94.67 \\ 
dict & suffix &  surface &\textbf{ 37.7} & 93.46 &\textbf{42.2} &93.23  \\
dict & factors & surface & 37.5 & \textbf{95.72} & 42.0 & \textbf{95.11} \\
\hline

\end{tabular}
\caption{Comparison of constraint integration methods on the oracle test set. All the models were trained on non-lemmatized, surface form constraints.}
\label{tab:integration}
\end{center}
\vspace{-0.6cm}
\end{table*}

\begin{table*}[h]
\begin{center}
\begin{tabular}{llccccc}
\textbf{Train prep.} & \textbf{Test prep.}   & \textbf{BLEU} & \textbf{Cvg} &\textbf{BLEU$_L$} & \textbf{Cvg$_L$ } & \textbf{Pos $\rho$}\\

\toprule
baseline & no constraints &   32&	68.84 & 38.2 & 78.14 & 0.9404\\

lemma & no constraints & 31.8 & 69.76 & 37.9 & 79.01 & 0.9367\\
lemma & lemma & \textbf{33.3} & 82.15 & \textbf{39.6} &\textbf{ 93.42} &\textbf{ 0.9341}\\

stemming & no constraints & 31.3	&69.51 & 37.4 & 78.48 & 0.9338\\
stemming & stemming &  33.2 & \textbf{84.10} & 39.5 &  92.86 & 0.9235 \\

\hline

\end{tabular}
\caption{Comparison of stemming and lematization as a preprocessing for training and test constraints.}
\label{tab:stem}
\end{center}
\vspace{-0.6cm}
\end{table*}

\subsection{Wiktionary vs. proprietary dictionary}
\label{app:wiki}

Dictionary is necessary for one of the training methods we explore. For our main results, we used a proprietary dictionary, which provides better coverage of the possible term pairs, but harms the reproducibility of this part of our experiments. Thus, we also evaluated our method using Wiktionary\footnote{\url{https://www.wiktionary.org/}} to obtain constraints in the same way as described in the main experiments section (see Section \ref{sec:experiments}). We present the results in Table \ref{tab:wiki}.

Looking up term pairs from the commercial dictionary in the test set, we found 7201 term pairs that were used as a constraint. On the other hand, we found only 2529 term pairs using Wiktionary.  We see that both models are able to incorporate constraints from the dictionary used during the training with similar success -- about 94\% of the constraints are covered. However, \textit{Large} dictionary provides better BLEU scores, since more constraint pairs are found overall in the test set.

\subsection{Mixed lemma and surface form training}

As we noticed in Section~\ref{sec:experiments}, lemmatized models have lower surface form coverage than non-lemmatized models when supplied with constraints in the reference surface form. As we show in our manual analysis, this is mostly an issue of automated evaluation based on comparison with reference, as the constraints are produced in correct form given the context of the output sentence produced by the model. Nevertheless, we experimented with a way to improve results of this automatic evaluation.

We trained another batch of models with 50\% of the constraints lemmatized and 50\% left in the surface forms.
Table~\ref{tab:europarl_whole_mixed} shows that this type of training improves integration of reference surface form constraints over the training where all constraints are lemmatized, while performance on lemmatized constraints does not decrease by a large margin.

\subsection{Stemming}

In Table \ref{tab:stem}, we compare stemming\footnote{\url{https://research.variancia.com/czech_stemmer/}} and lemmatization as the contraint preprocessing methods.
The models in the table are trained with suffix constraints.
The results are very similar, with stemming obtaining better results in terms of surface form coverage whereas lemmatization is better in lemma coverage.
\mn{is it fair to evaluate stemming by lemma-based metrics?}
Since in \cref{sec:analysis} we have shown that the difference between surface and lemma coverage for lemmatized model is caused by the automatic reference-based evaluation and not by real errors in the translation, we opted for lemmatization in the paper.

\subsection{Constraint integration using factors}
\label{app:factors}

We also present preliminary experiments with the factored model for constraining inspired by \citet{dinu-etal-2019-training}.
We use a two-factor model, where the first factor comprises of the input sequence of words.
For each source constraint in the input sequence, the translation tokens are inserted immediately after.
In the second factor, one of the following three label values is assigned to a corresponding input token:
\begin{itemize}
\itemsep0em 
    \item \texttt{0}: ordinary source token without constraint
    \item \texttt{SRC}: source side of a constraint
    \item \texttt{TGT}: translation of the constraint
\end{itemize}
For instance, consider the example in Table~\ref{tab:factors}.
Each word in the first factor has an associated label in the second factor according to its role in the translation.
The words \textit{plánováno} and \textit{obcích} are Czech constraints that must appear in the translation of the English sentence.
As a part of the target constraint, both words are labeled with the value \texttt{TGT} in the second factor.
The words \textit{planned} and \textit{municipalities} are English words representing the source part of the constraints, thus receiving the value \texttt{SRC}.
Words that are not constrained are labeled by \texttt{0} in the second factor.

The values of the second factor are copied over all subwords of the constraint sequence.
Embeddings of the values in both factors have the same dimensionality $d_{emb}$ and they are summed to obtain a complete embedding, which is used by the model.
For example, function $E_{sub}$ produces an embedding of an input subword and function $E_{f}$ produces an embedding of its label.
Final embedding of the word \textit{planned} in the above example is computed by the following formula:
$$E(planned_{SRC})=E_{sub}(planned)+E_f(SRC)$$

\cref{tab:integration} shows the comparison with the other integration methods on the oracle test set, similar to \cref{sec:oracle}.
We see factors provide the best coverage of the constraints. The factored approach makes use of alignment between source and target. This additional information probably helps with generating the correct constraints, but also complicates the preprocessing.
Since the differences are only minor and the goal of our paper is not to reach state-of-the-art results in constrained translation, we opted for the suffix-based approaches for simplicity.
Nevertheless, we note that factored approach is promising and we plan further research in this direction.
\jj{add position correlation for factors}
\jj{Add comparison of positional emedding shifting, prefix vs. sufix, stemming vs. lemmatization}




\end{document}